\documentclass[runningheads]{llncs}
\usepackage{graphicx}
\usepackage{booktabs}            
\usepackage{multirow}            
\usepackage{amsfonts}            
\usepackage{tabularx}
\usepackage{duckuments}          
\usepackage{amssymb}
\usepackage{amsmath}
\usepackage{subcaption}
\usepackage{bm}
\usepackage{hyperref}
\usepackage{wrapfig}
\begin{document}
\title{Equivariant Parameter Sharing for Porous Crystalline Materials}
%
%
\author{Marko Petkovi\'{c}\inst{1,2}\orcidID{0009-0000-4918-6027} \and
Pablo Romero Marimon\inst{1}\orcidID{0000-0001-9164-6541} \and
Vlado Menkovski\inst{1,2}\orcidID{0000-0001-5262-0605}
\and
Sofia Calero\inst{1,2}\orcidID{0000-0001-9535-057X}}
\authorrunning{M. Petkovi\'{c} et al.}
%
\institute{Eindhoven University of Technology, Eindhoven, The Netherlands \\
\email{\{m.petkovic1, v.menkovski, s.calero\}@tue.nl}\\
\email{p.romero.marimon@student.tue.nl}\\
\and
Eindhoven Artificial Intelligence Systems Institute, Het Kranenveld 12, Eindhoven, The Netherlands\\
}
\maketitle              
\begin{abstract}
Efficiently predicting properties of porous crystalline materials has great potential to accelerate the high throughput screening process for developing new materials, as simulations carried out using first principles model are often computationally expensive. To effectively make use of Deep Learning methods to model these materials, we need to utilize the symmetries present in the crystals, which are defined by their space group. Existing methods for crystal property prediction either have symmetry constraints that are too restrictive or only incorporate symmetries between unit cells. In addition, these models do not explicitly model the porous structure of the crystal. In this paper, we develop a model which incorporates the symmetries of the unit cell of a crystal in its architecture and explicitly models the porous structure. We evaluate our model by predicting the heat of adsorption of CO$_2$ for different configurations of the mordenite zeolite. Our results confirm that our method performs better than existing methods for crystal property prediction and that the inclusion of pores results in a more efficient model.

\end{abstract}
\section{Introduction}
Deep Learning has shown to be of great use in materials science, in tasks like property prediction and high-throughput screening of potential materials \cite{choudhary2022recent}.  In these workflows, many materials are first simulated using first principles methods, such as Density Functional Theory (DFT) and classical simulation, such as Molecular Dynamics (MD),  to find candidate materials to synthesize. However, these simulations are often computationally expensive and can take days or weeks to simulate a single new material. With Deep Learning, it is possible to accelerate the process of finding suitable materials, by developing data-driven surrogate models. These models scale significantly better than first principle simulators and allow for efficient search of the space of potential candidates \cite{stein2019progress}. Graph Neural Network (GNN) architectures are commonly used for modeling molecules and materials \cite{reiser2022graph} as these objects can effectively be represented as a graph. However, general-purpose GNNs are too restrictive as they incorporate only a part of the symmetries and periodicity present in crystal structures.

To overcome these limitations, for crystalline materials, multiple GNN architectures have recently been proposed \cite{chen2019graph,choudhary2021atomistic,kabaequivariant,schutt2017schnet,xie2018crystal,yan2022periodic} that accurately predict the properties of materials. These methods are specific extensions of general-purpose GNN that preserve the geometric structure of the crystal in their data representation. Despite preserving the geometric structure, none of the proposed models explicitly encode any information regarding pores, as the empty space does not lie on the data domain, and is thus not taken into account. Furthermore, they do not make use of the crystal symmetries in the material representation, since they are typically equivariant to a symmetry group larger than the space group of the crystal. We hypothesize that model architectures that do not explicitly model the porous structure of porous materials will struggle to infer the relevance of atom arrangements around pores for different properties. 

\begin{figure}[tb!]
     \hspace*{\fill} 
     \centering
     \begin{subfigure}[b]{0.413\textwidth}
         \centering
         \includegraphics[width=\textwidth]{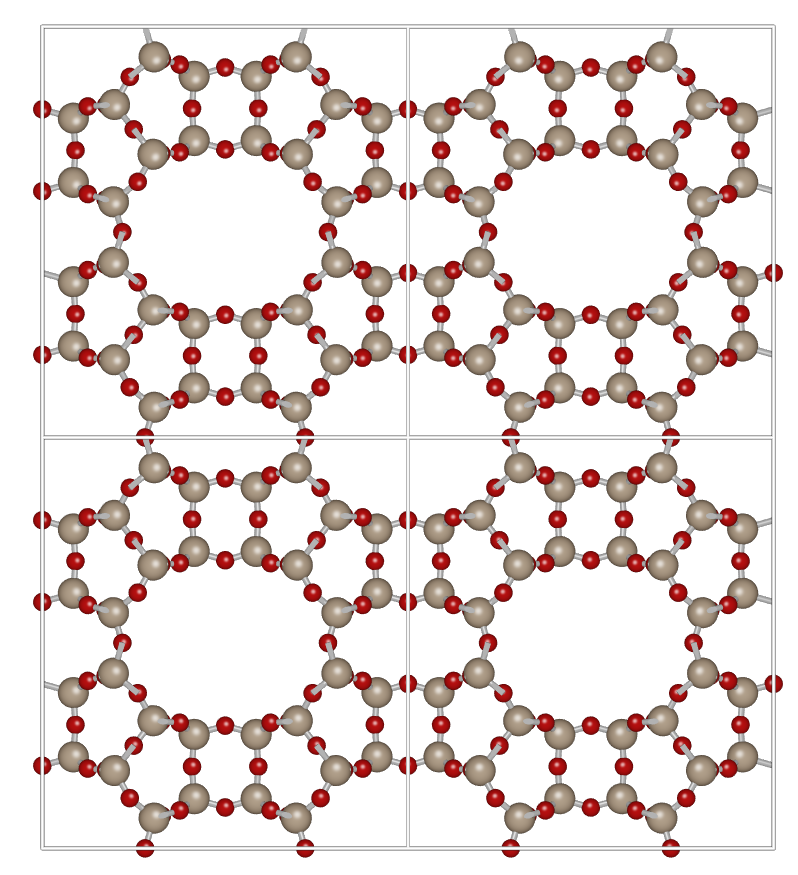}
         \caption{MOR}
         \label{fig:morunit}
     \end{subfigure}
     \hfill
     \begin{subfigure}[b]{0.3\textwidth}
         \centering
         \includegraphics[width=\textwidth]{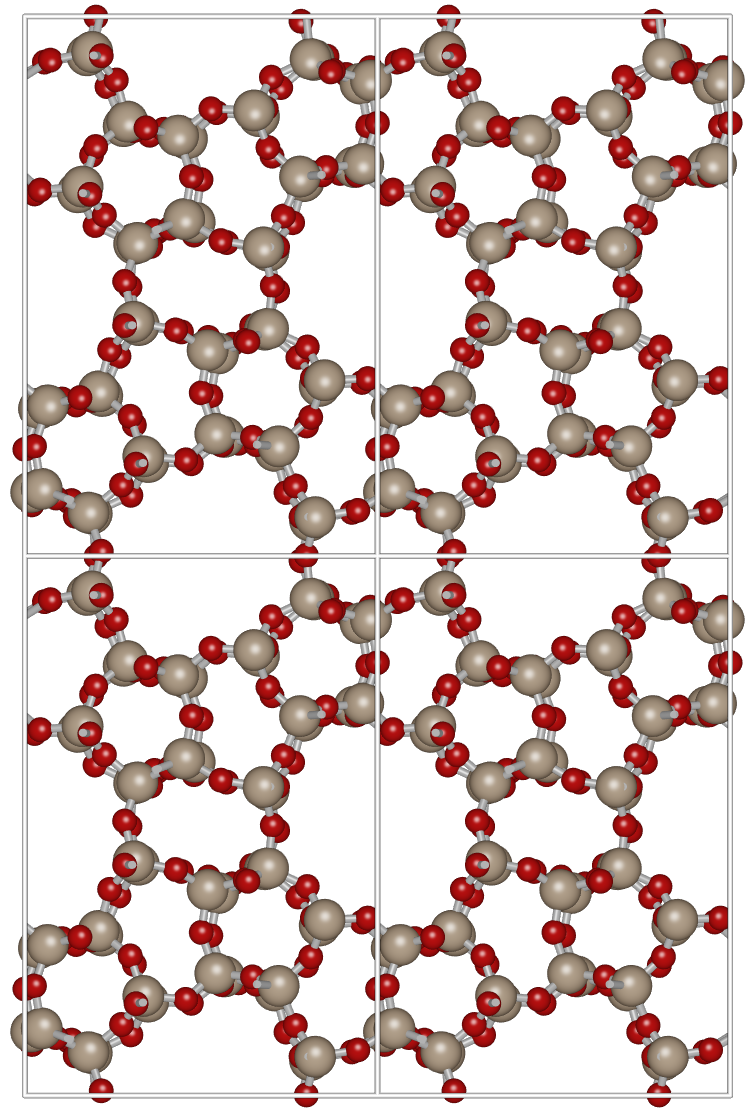}
         \caption{MFI}
         \label{fig:mfiunit}
     \end{subfigure}
     \hspace*{\fill} 
    \caption{Four unit cells of all silica MOR and MFI viewed along the z-axis and y-axis respectively. Images were generated using iRASPA \cite{dubbeldam2018iraspa}.}\label{fig:mormfi}
    \vspace{-5mm}
\end{figure}

Zeolites are a type of porous, crystalline materials of particular interest, as they are easily synthesizable \cite{khaleque2020zeolite}. They are used in applications such as gas separation and are a potential method for carbon capture \cite{sneddon2014potential}. The crystal structure of zeolites consists of TO$_4$ tetrahedra. In these tetrahedra, the T-atoms can either be aluminium or silicon, and both have different influences on the properties of the material. All four corners of the tetrahedra are shared, which results in a porous material. In Figure \ref{fig:morunit}, the porous structure of the Mordenite (MOR) and ZSM-5 (MFI) zeolites can be seen. The ability to capture CO$_2$ of a zeolite can be measured by its heat of adsorption in kJ/mol and is calculated as follows: $-\Delta H = \Delta U - RT$. Here, $\Delta U$ is the difference in internal energy before and after adsorption, $R$ is the universal gas constant, and $T$ is the temperature. The heat of adsorption can be influenced by the structure of the different types of zeolites and the amount and distribution of aluminium atoms in the framework \cite{choi2022effect,moradi2021effect,yang2018atomistic}.

Due to the difference in charge between aluminium and silicon atoms, it is necessary to balance the charge when aluminium atoms are present in a framework. To achieve this, cations such as sodium are inserted in the crystal structure. Since the cations are positively charged, they additionally attract CO$_2$ through Coulombic forces, thus increasing the ability of the material to adsorb CO$_2$. However, while the cations increase the adsorption strength, they also occupy physical space in the pores of the material. When multiple sodium cations are inside a pore, they can restrict CO$_2$ from entering it. As a result, the adsorption capacity of the material decreases. It is unclear which distributions of aluminium and silicon in different zeolites are optimal to maximize the heat of adsorption.

In this paper, we propose a novel GNN architecture that exploits information regarding the porous structure, as well as the symmetry of these materials, using parameter-sharing based on their space groups. This model allows us to effectively model the properties of porous crystalline materials.

We empirically validate our approach by modeling the heat of adsorption of CO$_2$ for the MOR and MFI zeolite. Our contributions are threefold: 1) We adapt the Equivariant Crystal Networks (ECN) architecture from \cite{kabaequivariant} to be equivariant with respect to the symmetry group of the unit cell. 2) We extend this architecture to explicitly model pores and show how this modification improves property prediction performance. 3) We introduce a new dataset containing different configurations of aluminium and silicon for the MOR and MFI zeolites, along with the CO$_2$ heat of adsorption values for the different configurations.

\section{Related Work}
\subsubsection{Machine Learning Methods for Crystals} Due to the success of different GNN architectures in modeling molecules, similar GNN architectures have been proposed for predicting material properties. Crystal Graph Convolutional Neural Networks (CGCNN) \cite{xie2018crystal} are one of the first architectures for crystals, which include periodicity in the data representation and are invariant with respect to permutations of atomic indices. In the MEGNet architecture \cite{chen2019graph}, a global state is used to improve the generalization of the model. Continuous filter convolutions have been introduced in the SchNet architecture \cite{schutt2017schnet}, which as a result can model the precise relative locations of atoms better when calculating local correlations. Another approach has been proposed in DimeNet \cite{gasteiger_dimenetpp_2020,gasteiger_dimenet_2020}, where the network also takes directional information between atoms into account. In ALIGNN \cite{choudhary2021atomistic}, the GNN processes simultaneously the graph and the line graph representation of the crystal, which takes the angles between edges into account. In addition, a transformer based architecture \cite{yan2022periodic} has been proposed, which additionally encodes the periodic nature of the crystal.  More recently a new approach has been proposed \cite{kabaequivariant}, which proposes a parameter-sharing scheme for message-passing. In this method, multiple unit cells are modelled, where parameters are shared based on the symmetry group of the crystal lattice. As such, the model gains in expressivity by encoding a part of the crystal symmetries in its architecture.
\subsubsection{Machine Learning in Porous Materials} Existing ML methods for porous materials frequently make use of feature engineering, which is used to predict properties by traditional ML models or shallow neural networks \cite{jablonka2020big,zhang2022accelerated}. Another approach made use of the CGCNN architecture \cite{wang2020accelerating} and extended it with engineered features \cite{wang2022combining} to improve performance. In their method, nodes in the graph representation do not correspond to atoms but rather correspond to secondary building units (SBUs), which consist of multiple atoms. 
\section{Crystal Symmetries}

\subsubsection{Unit Cell} Zeolites are crystalline materials, meaning that they contain an infinitely repeating pattern in all directions. This pattern be described by the set of integral combinations of linearly independent lattice basis vectors $\mathbf{a_i}$:
\begin{equation}
    \Lambda = \left\{ \sum_i^3 m_i \mathbf{a}_i\ |\ m_i \in \mathbb{Z}\right\}
\end{equation} 
The crystal lattice has an associated translation group $T_\Lambda$, which captures translational symmetry.  A unit cell is a subset of the lattice, which tiles the space of the crystal when translated by lattice vectors and is the minimum repeating pattern of the crystal. The unit cell is defined by the basis vectors as follows:
\begin{equation}
    U = \left\{ \sum_i^3 x_i \mathbf{a}_i\ |\  0 \leq x_i < 1 \right\}
\end{equation} 
The unit cell of a crystalline material contains a set of atomic positions, which is defined as $S = \{\mathbf{x}_i \; | \; \mathbf{x}_i \in U\}$, where $\mathbf{x}_i$ is the position of the atom in the unit cell. In addition to the set of atomic positions, we also define the set of pores contained in the unit cell. We define each pore using the atoms directly surrounding the pore. We represent each pore by the location of its center, as well as its surface area along which diffusion happens in terms of \AA$^2$. This results in the following set of pores: $P = \{(\mathbf{x}_{p_i}, area(p_i))\;|\; p_i \in U\}$, where $p_i$ is the pore and $\mathbf{x}_{p_i}$ is the centre of the pore.

\subsubsection{Space Groups} In crystalline materials, there are often multiple symmetries present inside the unit cell, defined by a space group $G$. The space group $G$ is the set of isometries that maps the crystal structure onto itself. Each element of the space group can be expressed as a linear transformation $\mathbf{W}$ and a translation $\mathbf{t}$, represented by a tuple ($\mathbf{W}$, $\mathbf{t}$). When mapping a vector $\mathbf{x}$ using an element of the space group, it is mapped to $\mathbf{Wx}$ + $\mathbf{t}$.

Inside a unit cell, every element of the space group $G$ maps the atomic/pore positions in the unit cell onto itself. While the type of atom at a certain position in the unit cell might change as a result of a transformation, the material remains the same. Therefore, each group action $g$ of the space group can be considered a permutation of the atoms and pores in the unit cell.  

\subsubsection{Group Orbits} The orbit of an element is created by applying all of the different elements of a space group $G$ to it. If the element is a vector $\mathbf{x}$, its orbit is the set of vectors to which the element can be moved by the group action. The orbit of $\mathbf{x}$ is defined as follows:
\begin{equation}
    G \cdot \mathbf{x} = \left\{ g \cdot \mathbf{x}\ |\ g \in G \right\}
\end{equation}

\section{Methods}

\subsubsection{Crystal Representation}
In our crystal representation, we only consider the set of atoms inside of the unit cell, as the content of a unit cell fully defines the porous structures and symmetry of the material. Each atom in the unit cell is represented by a feature vector $\mathbf{t}_i$ that is a one-hot encoding of the atom type. Next to this, we represent each pore inside the unit cell with a feature vector $\mathbf{p}_i$, which contains its surface area, as well as the number of atoms surrounding it. 

To represent the topology of the atoms and pores we construct a graph, where each atom is represented with a node. When the crystal contains clearly defined covalent bonds these can be used as edges in the graph, like in the case of zeolites. Pores are included in the graph representation by adding additional edges between the pore nodes and each atom on the boundary of the pore. By including these nodes, all atoms around the same pore are reachable from each other at most in two steps. Without pore nodes, this number could have been significantly larger, particularly for crystals with larger pores. 

The notion of a  pore has a certain analogy to the global feature vector introduced in MEGNet \cite{chen2019graph}. However, our approach is distributed in the geometry of the crystal which in turn allows the GNN-based model to learn locally distributed features, which is a more parameter-efficient solution. 

Based on \cite{schutt2017schnet}, we make use of radial basis functions to encode the distance between two neighboring nodes in the graph. We calculate the edge embedding $\mathbf{e}_{ij}$ as in Equation \ref{eq:rbf}, where $\gamma$ and $\boldsymbol{\mu}$ are hyperparameters. When calculating the distance between two atoms, we respect the periodic boundary conditions set by the unit cell by using the minimum image convention. Thus, we treat the opposite boundaries as a single boundary and consider the atoms and pores as neighbors and therefore sharing an edge.
\begin{equation}\label{eq:rbf}
    \mathbf{e}_{ij} = \exp \left(-\gamma(\|\mathbf{x}_i -\mathbf{x}_j\| - \boldsymbol{\mu})^2 \right)
\end{equation} 

Since we are developing a network architecture to predict properties based on the silicon and aluminium configuration in zeolites, we do not explicitly encode the oxygen atoms as nodes, as only the atoms placed in the T-sites of each TO$_4$ tetrahedron can change, while oxygen atoms always remain in the same position. 

\subsubsection{Equivariant Message Passing}
Since the space group acts as a permutation on the atoms and pores in the unit cell, we can describe the action of a group element using Equation \ref{eq:groupact_perm}. Here, $\pi_g^\mathbf{t}$ and $\pi_g^\mathbf{p}$ are the permutations of the atoms and pores as a result of group action $g$.  
\begin{equation} \label{eq:groupact_perm}
    \left( g\mathbf{t}_i = \mathbf{t}_{\pi_g^\mathbf{t}(i)} \wedge g\mathbf{p}_j = \mathbf{p}_{\pi_g^\mathbf{p}(j)} \right) \forall g \in G
\end{equation} 
As we model different configurations of the same crystal structure using our architecture, the model needs to be equivariant to $G$. The defined model is based on the message passing framework \cite{gilmer2017neural}, which we extend by defining parameter-sharing patterns \cite{ravanbakhsh2017equivariance} for the message and node update functions. 
\begin{figure}[t!]
\hspace*{\fill}
     \centering
     \begin{subfigure}[b]{0.35\textwidth}
         \centering
         \includegraphics[width=\textwidth]{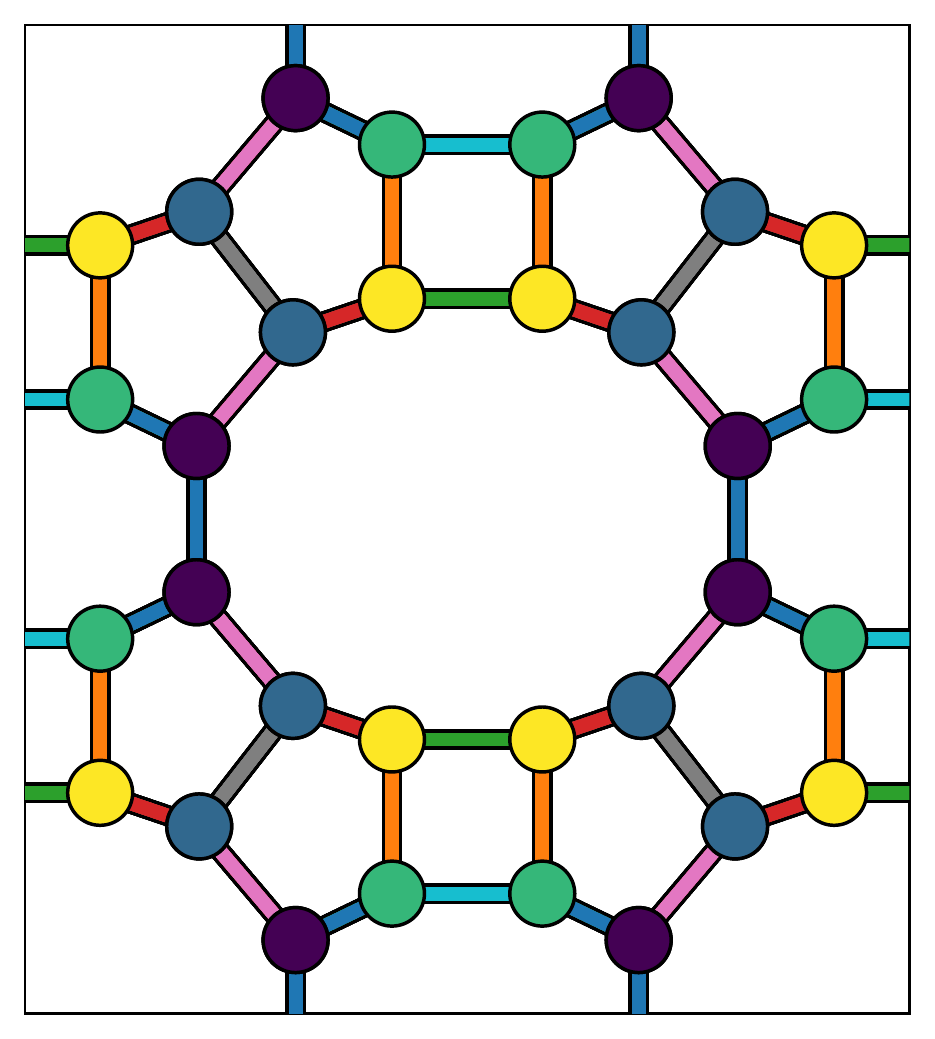}
         \caption{Without pores}
         \label{fig:noporews}
     \end{subfigure}
     \hfill
     \begin{subfigure}[b]{0.35\textwidth}
         \centering
         \includegraphics[width=\textwidth]{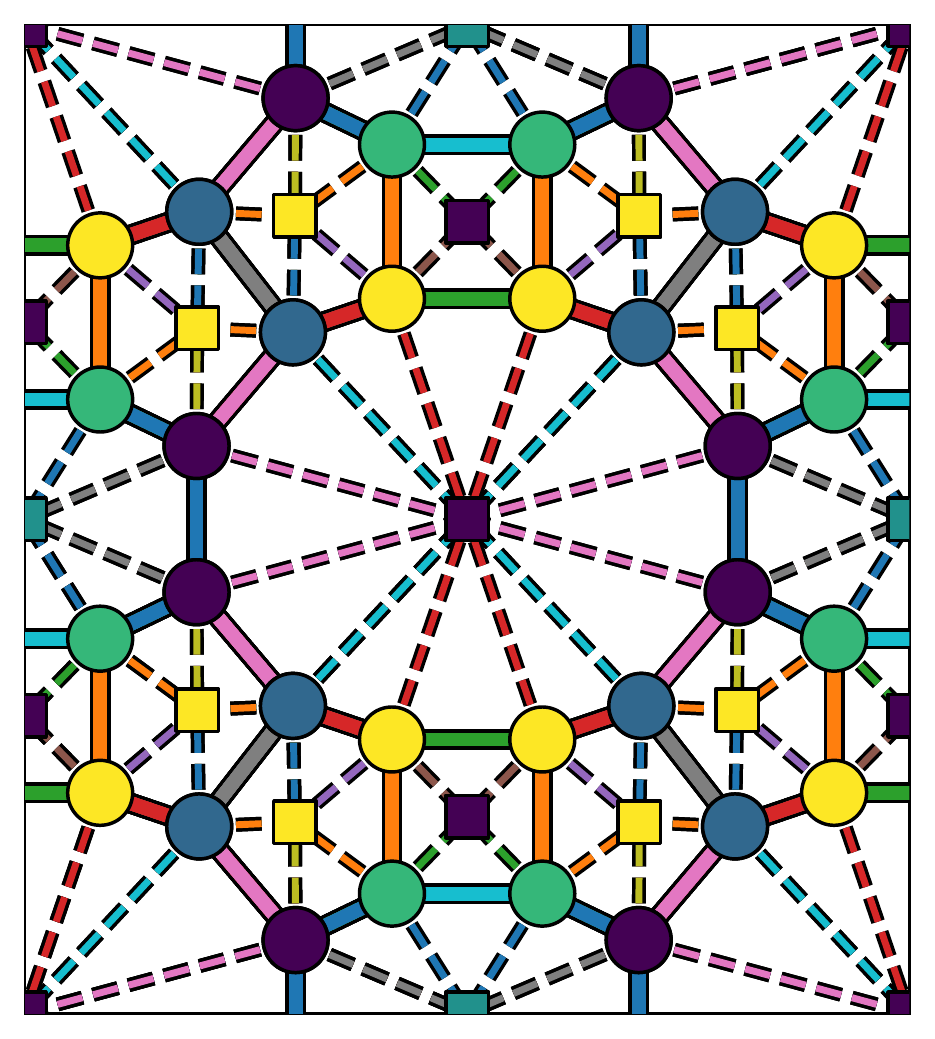}
         \caption{With pores}
         \label{fig:porews}
    
     \end{subfigure}
     \hspace*{\fill}
    \caption{Weight sharing scheme for MOR (z-axis). Nodes/edges of the same color share parameters in their node/edge update functions. Dashed edges are between atoms (circle) and pores (squares). Solid edges are between atoms. 
    }
    \label{fig:weightsharing}
    \vspace{-5mm}
\end{figure}

First, we define how a parameter-sharing pattern for a graph is calculated. Following \cite{kabaequivariant}, we define the parameter-sharing pattern as the colored bipartite graph $\mathrm{\Omega} \equiv (\mathbb{N}, \alpha, \beta)$. Here, $\mathbb{N}$ is the set of input atoms and pores, $\alpha$ is the edge color function ($\alpha: \mathbb{N} \times \mathbb{N} \rightarrow \{1,...,C_e\}$) and $\beta$ is the node color function ($\beta: \mathbb{N} \rightarrow \{1,...,C_h\} $). $C_e$ and $C_h$ are the amounts of unique edge and node colors respectively. As shown in Equations \ref{eq:edgecolor} and \ref{eq:nodecolor}, the color functions take the same value if two edges ($(i,j), (k,l)$) or atoms/pores ($i,j$) lie on each other's orbit. 
\begin{align}
    \alpha(i,j) = \alpha(k,l) &\iff (k,l) \in G \cdot (i,j) \label{eq:edgecolor}\\
    \beta(i) = \beta(j) &\iff j \in G \cdot i \label{eq:nodecolor}
\end{align} When introducing a parameter-sharing pattern based on the edge and node coloring function, we are effectively introducing an additional message (node) update function for each unique message (node) in the graph representation of the crystal. Following the proof of Claim 6.1 from \cite{kabaequivariant}, the model architecture remains equivariant to the space group of the crystal. In Figure \ref{fig:weightsharing}, the parameter-sharing pattern for MOR can be found, where nodes and edges are colored according to Equations \ref{eq:edgecolor} and \ref{eq:nodecolor}. The message passing operation equivariant to the space group is defined in Equations \ref{eq:poregnn1}-\ref{eq:poregnn3}, where $\mathbf{t}_i$ is the atom embedding and $\mathbf{p}_i$ is the pore embedding. Here, superscript $h$ indicates messages between atoms, $k$ messages from pores to atoms and $l$ messages from atoms to pores.

Since different types of crystals have different amounts of atoms and/or different space groups, we cannot share the parameters of message-passing operations between crystals. As a result, each crystal topology requires its own model. 
\small
\begin{align}
        &\mathbf{m}_{ij}^h = \phi_e^{\alpha(i,j)}(\mathbf{t}_i^t, \mathbf{t}_j^t, \mathbf{e}_{ij}),   & &\mathbf{m}_{ij}^k = \phi_e^{\alpha(i,j)}(\mathbf{t}_i^t, \mathbf{p}_j^t, \mathbf{e}_{ij}), &  &\mathbf{m}_{ij}^l = \phi_e^{\alpha(i,j)}(\mathbf{p}_i^t, \mathbf{t}_j^t, \mathbf{e}_{ij}), \label{eq:poregnn1}\\
    &\mathbf{m}_{i}^h = \frac{1}{|N_i^h|} \sum_{j \in N_i^h} \mathbf{m}_{ij}^h, & &\mathbf{m}_{i}^k = \frac{1}{|N_i^k|} \sum_{j \in N_i^k} \mathbf{m}_{ij}^k, & &\mathbf{m}_{i}^l = \frac{1}{|N_i^l|} \sum_{j \in N_i^l} \mathbf{m}_{ij}^l, \label{eq:poregnn2}\\
    &\mathbf{t}_i^{t+1} = \phi_h^{\beta(i)}(\mathbf{t}_i^t, \mathbf{m}_i^h, \mathbf{m}_i^k), & &\mathbf{p}_i^{t+1} = \phi_h^{\beta(i)}(\mathbf{p}_i^t, \mathbf{m}_i^l).\label{eq:poregnn3}
\end{align} 
\normalsize

\begin{figure}[t!]
    \centering
    \includegraphics[width=\textwidth]{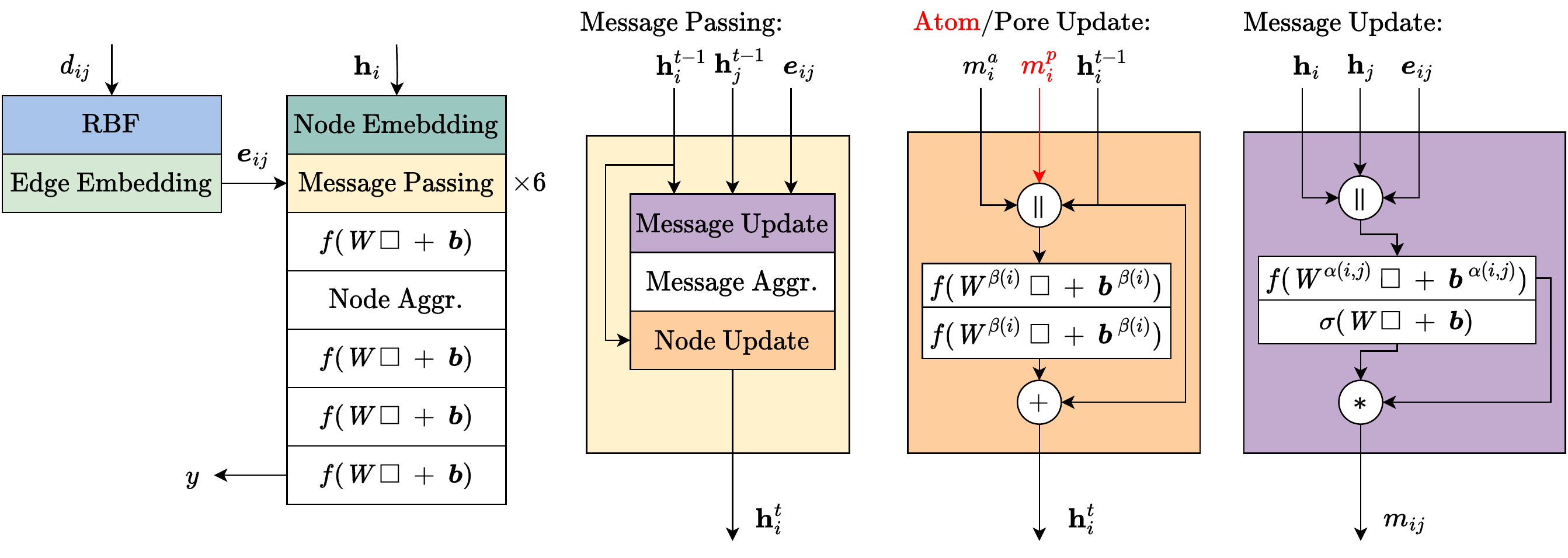}
    \caption{Overview of the network architecture with pores. The $\square$ and $||$ denote the layer's input and concatenation respectively. The $+$ and $*$ denote elementwise summation and multiplication. $\mathbf{h}_i$ represents the embedding of node $i$ (which can be an atom or a pore), while $d_{ij}$ represents the distance between nodes $i$ and $j$. $m_i^a$ is the aggregated message received from atoms and $m_i^p$ from pores. $f$ and $\sigma$ represent the leaky ReLU and sigmoid activation function respectively. $(\bm{W/b})^{\alpha(i,j)}$ and $(\bm{W/b})^{\beta(i)}$ denote the set of weights for the value of coloring functions $\alpha(i,j$) and $\beta(i)$. In the node update function, pore nodes do not receive messages from pores ($m_i^p$, red). For all aggregations, sum-pooling is used. Finally, the node embedding operation is different for atom and pore nodes.}
    \label{fig:modelarch}
    \vspace{-6.2mm}
\end{figure}

\section{Experiments}
\subsubsection{Network Architecture} 
In Figure \ref{fig:modelarch}, an overview of the model architecture is presented. First, the edges are embedded using Equation \ref{eq:rbf} on their distance ($d_{ij}$), which is followed by a fully connected layer. Simultaneously, both atoms and pores are embedded using two different fully connected layers, as atoms have only one feature while pores have two. In the message update function, the embeddings of the sending and receiving node and the edge embedding are concatenated, which is followed by a fully connected layer that shares weights according to Equation \ref{eq:edgecolor}. Each message also receives a weighting factor, calculated by the fully connected layer following the equivariant layer. 

The message aggregation step depends on the node type. Because pore nodes are included in the graph, we may distinguish between two types of messages based on whether they are sent from an atom or a pore. To distinguish between the origin of messages, we separately aggregate messages sent from pores ($m_i^p$) and atoms ($m_i^a$). Following the concatenation of the different messages and the node embedding, we apply 2 linear layers. The node update block also contains a residual connection \cite{he2016deep}.

Following multiple message-passing steps, the node features are sum-aggregated, and an MLP is applied to obtain the final prediction. In the model with pores, we only aggregate the pore node features. This way, we implicitly force the model to learn the contribution of each pore to the adsorption capacity.

\begin{wrapfigure}{r}{0.425\textwidth}
\vspace{-8.5mm}
 \centering
\includegraphics[width=0.4\textwidth]{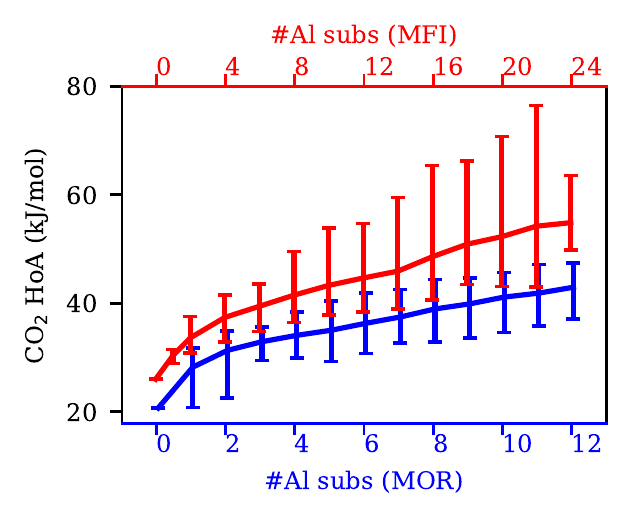}
\
 \caption{Heat of Adsorption distributions per amount of Al atoms.}
\label{fig:co2dist}
\vspace{-8mm}
\end{wrapfigure}

As we are training the model on a single crystal structure at a time, it suffices to keep the architecture relatively simple. We perform 6 message passing steps, with internal hidden states of size 16. Then, we process each atom (pore) with an MLP with an output size of 24. Following this, sum-aggregation is performed, after which the final MLP makes the prediction. Our model implementation is based on ECN \cite{kabaequivariant} and uses the PyTorch \cite{paszke2019pytorch} and PyTorch scatter \cite{torchscatter} packages as well as the AutoEquiv library \cite{autoequiv2021}. The main difference with ECN is that our architecture only models one unit cell and the symmetries within it, while ECN models multiple unit cell and the symmetries between the them.

\subsubsection{Data Generation}

To generate a dataset with porous, crystalline materials, we made use of the Mordenite and ZSM-5 zeolite frameworks. MOR contains 48 T-atoms, and adsorbates diffuse along the z-axis. MFI is a more complex zeolite, containing 96 T-atoms. It has multiple intersecting pores, where adsorbates can diffuse along the x-axis and y-axis. As a result, the adsorption capacity between two configurations with an equal amount of aluminium atoms can vary greatly. 

We generated multiple MOR and MFI frameworks with varying aluminium and silicon distributions, and carried out simulations to calculate the corresponding CO$_2$ heat of adsorption for each configuration. For generating the structures, the Zeoran program \cite{romero2023adsorption} was used and atom coordinates where taken from \cite{baerlocher2007atlas}. For MOR, 4992 structures were generated, where each structure contains up to 12 aluminium atoms. For MFI, 3296 structures were generated with up to 24 aluminium atoms. The amount of aluminium atoms was chosen such that they roughly match what is possible in practice. The program makes use of four different algorithms, which places aluminium atoms in either random positions, chains, clusters or homogeneously throughout the zeolite framework. To calculate the CO$_2$ heat of adsorption Grand-Canonical Monte Carlo simulations were carried out using the Widom Particle Insertion method \cite{widom1963some}, performed with the RASPA software \cite{dubbeldam2016raspa}. Frameworks were considered rigid and the force field parameters for the interactions between the zeolite and the adsorbate were taken from \cite{garcia2009transferable}. The force field for carbon dioxide was taken from \cite{harris1995carbon}.

In Figure \ref{fig:co2dist}, the 95\% confidence interval of the heat of adsorption per amount of aluminium atoms is shown. While there is a strong correlation between the amount of aluminium atoms and the heat of adsorption, there is still significant variance in the heat of adsorption for each amount of aluminium substitutions. 

\footnotesize
\begin{wraptable}{r}{0.33\textwidth}
\vspace{-8.35mm}
  \centering
    \caption{Parameter count.}
  \begin{tabular}{lc}
    \toprule
     Model & Parameters \\
    \midrule
    CGCNN & 0.11M \\
    MEGNet & 0.19M \\
    Matformer & 2.77M \\
    DimeNet++ & 1.74M \\
    SchNet & 0.44M \\
    ALIGNN & 4.01M \\
    ECN & 2.81M \\
    Ours (MOR) & 0.03M \\
    Ours (MFI) & 0.15M \\
    \bottomrule
  \end{tabular}
  \vspace{-5.5mm}
  \label{tab:params}
\end{wraptable}
\normalsize
\subsubsection{Model Evaluation} To evaluate the predictive performance of our models, we made an uninformed, random assignment of samples to the training (90\%) and testing set (10\%). We compare the performance of our model to different baselines \cite{chen2019graph,choudhary2021atomistic,gasteiger_dimenetpp_2020,kabaequivariant,schutt2017schnet,xie2018crystal,yan2022periodic}. For each baseline, we use the hyperparameters from their original papers. In addition, we conducted an ablation study, where we excluded pores and symmetries from our model to asses their contribution.  

Since predicting the heat of adsorption is a regression task, we made use of the Huber loss function. We used the AdamW optimizer \cite{loshchilov2017decoupled} with a learning rate of 0.001, and trained each model for 200 epochs. We report the mean-absolute error (MAE) and mean-squared error (MSE). To obtain confidence bounds, we trained each model 10 times with random weight initialization. The code for the model implementations and the zeolite dataset are available on \url{www.github.com/marko-petkovic/porousequivariantnetworks}.


\footnotesize
\begin{table}[t!]
    \centering
    \caption{Performance of different model architectures on CO$_2$ heat of adsorption prediction for the MOR and MFI datasets.}
    \begin{tabular}{lcccc}
\toprule
{} &  \multicolumn{2}{c}{MOR} & \multicolumn{2}{c}{MFI}\\
\midrule
{} &                  MAE &                  MSE  & MAE & MSE\\
\midrule
CGCNN & 1.374 $\pm$ 0.033 & 3.414 $\pm$ 0.107  & 2.814 $\pm$ 0.047 & 18.949 $\pm$ 0.561 \\
MEGNet & 1.260 $\pm$ 0.086 & 2.785 $\pm$ 0.290 & 2.674 $\pm$ 0.040 & 16.533 $\pm$ 0.701 \\
Matformer & 1.002 $\pm$ 0.074 & 1.843 $\pm$ 0.237 & 2.577 $\pm$ 0.372 & 12.552 $\pm$ 2.545\\
DimeNet++ & 0.938 $\pm$ 0.028 & 1.568 $\pm$ 0.076 & 2.862 $\pm$ 0.027 & 19.344 $\pm$ 0.472 \\
SchNet & 0.895 $\pm$ 0.016 & 1.482 $\pm$ 0.055 & 1.876 $\pm$ 0.047 & $\mathbf{6.826 \pm 0.270}$ \\
ALIGNN & 0.828 $\pm$ 0.035 & 1.293 $\pm$ 0.102 & $\mathbf{1.819 \pm 0.033}$ & 6.840 $\pm$ 0.187\\
\midrule
ECN & 1.282 $\pm$ 0.028 & 2.984 $\pm$ 0.124 & 2.484 $\pm$ 0.046 & 12.942 $\pm$ 0.610 \\
Ours (w/o pores/syms) & 1.184 $\pm$ 0.048 & 2.503 $\pm$ 0.163   & 2.717 $\pm$ 0.063 & 16.775 $\pm$ 1.093 \\
Ours (w/o syms) & 0.901 $\pm$ 0.040 & 1.505 $\pm$ 0.094   & 2.303 $\pm$ 0.110 & 11.841 $\pm$ 1.338 \\
Ours (w/o pores) & 0.904 $\pm$ 0.023 & 1.546 $\pm$ 0.075   & 2.029 $\pm$ 0.058 & 8.777 $\pm$ 0.375 \\
Ours & $\mathbf{0.813 \pm 0.010}$ & $\mathbf{1.286 \pm 0.038}$  & 1.902 $\pm$ 0.024 & 8.184 $\pm$ 0.288 \\
\bottomrule
\end{tabular}
    \label{tab:results}
    \vspace{-6.5mm}
\end{table}
\normalsize

As can be seen in Table \ref{tab:results}, our model obtains the best results for MOR, and achieves competitive results with SchNet and ALIGNN on MFI, despite using significantly fewer parameters (Table \ref{tab:params}). In addition, our model outperformed the ablated versions. For MFI, the other baselines achieve significantly worse results. We speculate that this behaviour is caused by the spatial features of the zeolites not explicitly being encoded in the graph representation. 

\begin{wrapfigure}{r}{0.4\textwidth}
\vspace{-8mm}
 \centering
\includegraphics[width=0.38\textwidth]{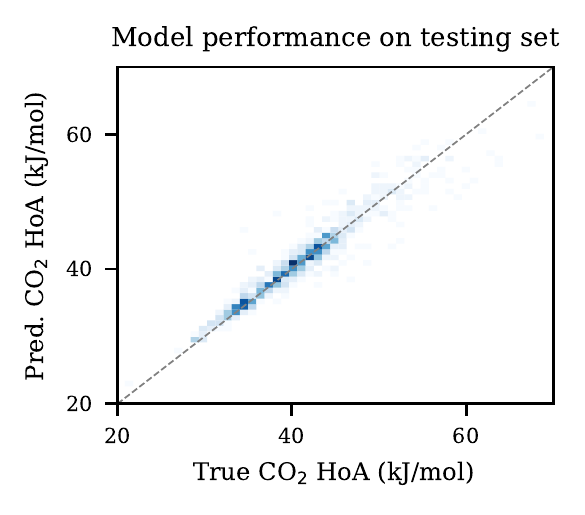}
\caption{Predicted against true heat of adsorption.}
\vspace{-8mm}
\label{fig:truepred}
\end{wrapfigure}

In Figure \ref{fig:truepred}, we the true against the predicted heat of adsorption values for our best model with pores on the MOR and MFI datasets are shown. We see that most predictions for both models are accurate. For higher heat of adsorption values, the models  perform slightly worse. This may be due to an insufficient amount of training examples present with a high heat of adsorption.

\begin{figure}[h] 
\vspace{-5mm}
     \hspace*{\fill} 
     \centering
     \begin{subfigure}[b]{0.4\textwidth}
         \centering
         \includegraphics[width=\textwidth]{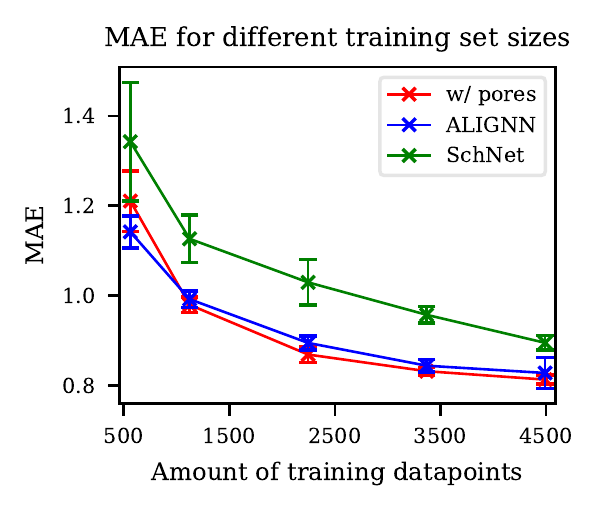}
         \caption{MOR}
         \label{fig:mae_mor}
     \end{subfigure}
     \hfill
     \begin{subfigure}[b]{0.4\textwidth}
         \centering
         \includegraphics[width=\textwidth]{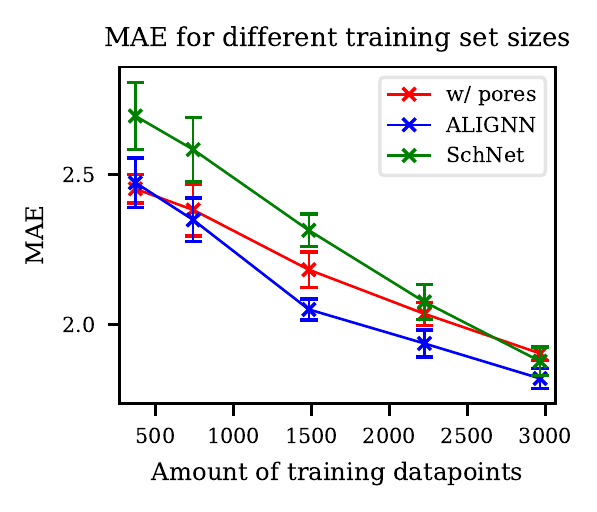}
         \caption{MFI}
         \label{fig:mae_mfi}
     \end{subfigure}
     \hspace*{\fill} 
     \caption{Data efficiency (MAE on test set) with different amounts of training data.}\label{fig:porevsequi}
     \vspace{-8mm}
\end{figure}

In addition, we carried out experiments to compare the data efficiency of our model, ALIGNN and SchNet. Here, we trained each model using different fractions ($\frac{1}{8}, \frac{1}{4}, \frac{1}{2}, \frac{3}{4}$ and 1) of the training set, and evaluated them on the same testing set. In Figure \ref{fig:porevsequi}, we see that our model achieves a data efficiency comparable to ALIGNN and better than SchNet for MOR. In the case of MFI, we see that our model has a slightly lower data efficiency than ALIGNN, while performing slightly better than SchNet for low amounts of training data. 

\section{Discussion}
We have proposed a new type of network which can exploit both symmetries inside the unit cell as well as the structure of porous crystalline materials. Our method achieved excellent performance on the CO$_2$ heat of adsorption prediction task, and has also shown a better parameter efficiency and a competitive data efficiency. This class of models has a significant potential to accelerate the high throughput screening of porous materials, by quickly narrowing the search space for candidate materials. 

%
%
%
\bibliographystyle{splncs04}
\bibliography{mybibliography}
\end{document}